\documentclass[preprint,12p]{elsarticle}
\makeatletter
\def\ps@pprintTitle{%
  \let\@oddhead\@empty
  \let\@evenhead\@empty
  \let\@oddfoot\@empty
  \let\@evenfoot\@oddfoot
}
\makeatother
\usepackage{subcaption}
\usepackage{float}
\usepackage{adjustbox}
\usepackage{multirow}
\usepackage{colortbl}
\usepackage{geometry}

\newlength{\twosubht}
\newsavebox{\twosubbox}

\usepackage{amssymb}
\usepackage{amsmath}

\begin{document}
\begin{frontmatter}

\title{Evaluating Artificial Intelligence Algorithms for the Standardization of Transtibial Prosthetic Socket Shape Design}

\author[label1]{C.H.E. Jordaan\corref{cor1}}
\ead{celena.jordaan@radboudumc.nl}
\author[label1]{M. van der Stelt}
\author[label1]{T.J.J. Maal}
\author[label2,label6]{V.M.A. Stirler}
\author[label3]{R. Leijendekkers}
\author[label4]{T. Kachman}
\author[label1]{G.A. de Jong}

\affiliation[label1]{organization={3D Lab, Radboud University Medical Centre},
            addressline={Geert Grooteplein Zuid 10}, 
            city={Nijmegen},
            postcode={6525 GA}, 
            state={Gelderland},
            country={The Netherlands}}
\affiliation[label2]{organization={Department of Trauma Surgery, Radboud University Medical Centre},
            addressline={Geert Grooteplein Zuid 10}, 
            city={Nijmegen},
            postcode={6525 GA}, 
            state={Gelderland},
            country={The Netherlands}}
\affiliation[label3]{organization={Department of Rehabilitation, Radboud University Medical Centre},
            addressline={Geert Grooteplein Zuid 10}, 
            city={Nijmegen},
            postcode={6525 GA}, 
            state={Gelderland},
            country={The Netherlands}}
\affiliation[label4]{organization={Donders Centre for Cognition, Radboud University},
            addressline={Houtlaan 4}, 
            city={Nijmegen},
            postcode={6525 XZ}, 
            state={Gelderland},
            country={The Netherlands}}
\affiliation[label6]{organization={Military Health Organisation, Ministry of Defence, Kromhout Kazerne},
            addressline={Herculeslaan 1}, 
            city={Utrecht},
            postcode={3584 AB}, 
            state={Utrecht},
            country={The Netherlands}}
            
\cortext[cor1]{Corresponding author}

\newgeometry{left=3cm,bottom=0.1cm}
\begin{abstract}
The quality of a transtibial prosthetic socket depends on the prosthetist's skills and expertise, as the fitting is performed manually. This study investigates multiple artificial intelligence (AI) approaches to help standardize transtibial prosthetic socket design. Data from 118 patients were collected by prosthetists working in the Dutch healthcare system. This data consists of a three-dimensional (3D) scan of the residual limb and a corresponding 3D model of the prosthetist-designed socket. Multiple data pre-processing steps are performed for alignment, standardization and optionally compression using Morphable Models and Principal Component Analysis. Afterward, three different algorithms - a 3D neural network, Feedforward neural network, and random forest - are developed to either predict 1) the final socket shape or 2) the adaptations performed by a prosthetist to predict the socket shape based on the 3D scan of the residual limb.  Each algorithm's performance was evaluated by comparing the prosthetist-designed socket with the AI-generated socket, using two metrics in combination with the error location. First, we measure the surface-to-surface distance to assess the overall surface error between the AI-generated socket and the prosthetist-designed socket. Second, distance maps between the AI-generated and prosthetist sockets are utilized to analyze the error's location. For all algorithms, estimating the required adaptations outperformed direct prediction of the final socket shape. The random forest model applied to adaptation prediction yields the lowest error with a median surface-to-surface distance of 1.24 millimeters, a first quartile of 1.03 millimeters, and a third quartile of 1.54 millimeters. 
\end{abstract}


\end{frontmatter}
\restoregeometry

\section{Introduction}
The World Health Organisation estimates that only one in ten people has access to assistive products, such as prostheses \citep{health-product-policy-and-standards-2017}. This problem is even more prevalent for the 30-40 million people with amputations in low and middle-income countries (LMICs), where only 5\% of people with amputations have access to assistive products such as prosthetic devices \citep{limbs_international}. In addition to limited availability, research by \citet{jensen_2006} indicated that only 58\% of sockets created in low-income countries provided an adequate fit, highlighting the need for quality improvements.  

Prosthetic sockets are traditionally made using: 1) the plaster cast method, where a prosthetist forms a negative mold of the patient's residual limb and modifies it by adding or removing plaster to achieve the desired socket shape, or 2) computer-aided design, in which a 3D scan of the residual limb is processed in specialized software to create the socket shape. A transtibial prosthetic socket mostly applies pressure to areas with more soft tissue, such as the back of the calf, while releasing pressure on bony areas, such as the end of the tibia \citep{safari_2015}. This commonly used method is the patellar tendon bearing (PTB) technique \citep{radcliffe_1961}. Another approach is the total surface bearing (TSB) technique, where the pressure is more equally distributed across the residual limb \citep{staats_1987}. The specific location and amount of pressure applied or reduced is patient-specific, making conventionally made prosthetic sockets highly dependent on the skills and expertise of a prosthetist \citep{safari_2020}. 

\begin{figure}[H]
    \centering
    \includegraphics[width=0.35\linewidth]{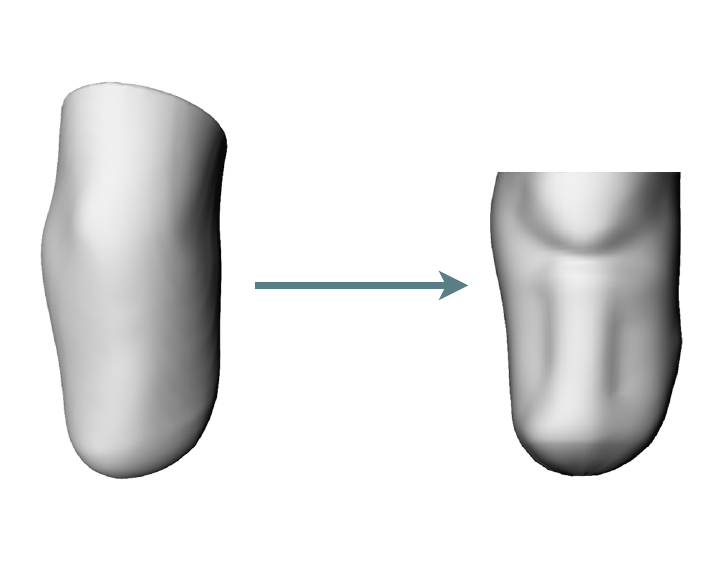}
    \caption{Example of the used data. A 3D scan of a stump (left) and the corresponding 3D model of the socket (right)}
    \label{fig:example_scan_stump_data}
\end{figure}

This study investigates multiple artificial intelligence (AI) algorithms to standardize transtibial socket shape design, aiming to increase the availability and quality of prosthetic care in LMICs. These algorithms are developed using data from prosthetists in the Dutch healthcare system. The data consists of a 3D scan of the residual limb and a corresponding 3D model of the interior wall of the prosthetic socket, crafted by a prosthetist.  Figure \ref{fig:example_scan_stump_data} shows an example of a 3D scan and the corresponding 3D socket shape model.

\section{Related work}
Several studies investigated the possibility of standardizing prosthetic socket fit. Research by \citet{colombo_2016} illustrated using a neural network to create a model that predicts the shape of a transtibial socket by exploiting a 3D scan of the stump, the patient's weight, lifestyle, and the tonicity of the residual limb. The algorithm predicts the corners of load and off-load areas of the stump: the patella, the patella tendon, and the tibia crest. Based on the patient's characteristics, the overall scaling of the socket and identified areas is performed. The results show that the corner identification of the load and off-load areas of the stump have a mean squared error of 3.8 millimeters. 

A prosthetic socket rectification model, researched by \citet{li_2019}, investigated a quantitative design using an eigenvector algorithm, based on ten independent prosthetists and stump characteristics, that indicates the required adaptations for transtibial sockets. Their method specified a numerical range of adaptations needed to create a socket. Therefore, it still requires a prosthetist's interpretation for accurate application. 
   
\citet{dickinson_2021} analyzed the variation in the shape of a residual limb and the shape of the socket using a combination of principal component analysis (PCA) and K-means clustering. Using PCA, they approximated the limb shape and socket design with 95\% confidence. Additionally, they created a PCA model on the combined limb shape and socket design, indicating the performed socket adaptations. However, this research did not provide a standardized way to fit transtibial prosthetic sockets. 

In research by \citet{cutti_2024} the creation of a prosthetist-specific rectification template was investigated. They collected data from 15 patients receiving a transtibial prosthetic socket and performed PCA on the adaptations between the residual limb and prosthetic socket of 14 patients. They applied the  PCA rectifications to the remaining data and found a mean radial error of 1.28 millimeters.  

This study continues on research performed by \citet{van_der_stelt_2024}, which developed a transtibial prosthetic socket shape design algorithm based on data from 116 patients. Similar to this research, the input for the developed algorithm is a 3D scan of the patient's stump, and the desired output is the corresponding 3D model of the socket. Using DiffusionNet, they predict the socket shape on the test set with a deviation of 2.51 millimeters \citep{sharp_2022}. Using AI on 3D data has several implications for handling this data, explained in the following section.  

\section{Theoretical background} 
\begin{figure}[!ht]
    \centering
    \begin{subfigure}[t]{0.148\textwidth}
        \centering
        \includegraphics[height=3cm, width = 2cm]{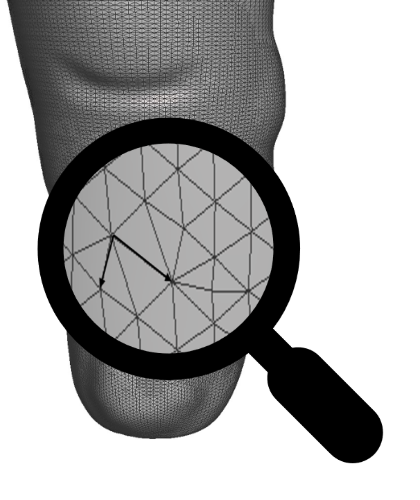}
        \caption{Point clouds are unstructured: the distances between data points are not fixed}
        \label{fig:pointcloud_unstructured}
    \end{subfigure}
    \hfill
    \begin{subfigure}[t]{0.401\textwidth}
        \centering
        \includegraphics[height=3cm, width = 5cm]{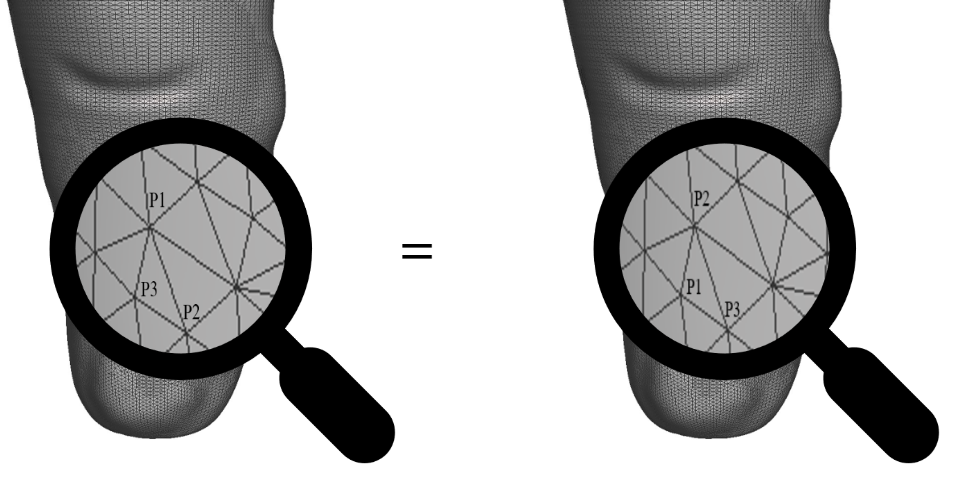}
        \caption{Point clouds are invariant: the order of the data points does not influence the final representation (as long as the winding order is equal)}
        \label{fig:pointcloud_invariant}
    \end{subfigure}
    \hfill
    \begin{subfigure}[t]{0.32998\textwidth}
        \centering
        \includegraphics[height=3cm]{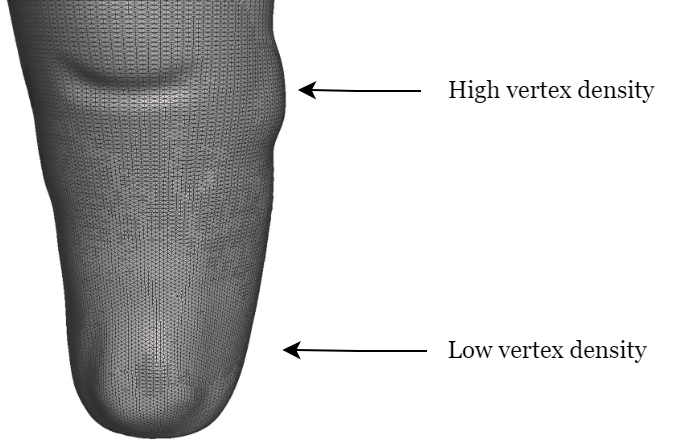}
        \caption{Point clouds are irregular: the number of data points can vary per location and between two point clouds representing the same object}
        \label{fig:pointcloud_irregular}
    \end{subfigure}
    \caption[Main challenges of 3D Artificial Intelligence]{Three main challenges of applying artificial intelligence on point clouds/meshes}
    \label{fig:3dai_challenges}
\end{figure}

As mentioned, this study uses 3D data,  represented as 3D meshes, composed of point clouds and interconnected faces. Applying AI techniques to 3D meshes presents several challenges. 3D meshes can have irregular densities, resulting in some areas having a higher concentration of points than others \citep{li_2018, qi_2017} (Figure \ref{fig:pointcloud_irregular}), which automatically results in unstructured point clouds, where points are unevenly distributed (Figure \ref{fig:pointcloud_unstructured}). Furthermore, 3D meshes are in some cases labeling invariant, meaning that changing the labeling of points still results in the same shape \citep{li_2018} (Figure \ref{fig:pointcloud_invariant}). Due to these challenges, more conventional techniques, such as convolutional neural networks, cannot be applied directly on 3D meshes. 

To deal with the stated challenges, the point clouds of the 3D models are standardized using a morphable model. Standardization through resampling is performed using the MeshMonk morphable model, which ensures a uniform and ordered sampling of the data \citep{white_2019}.  Following, three different AI methods are used to investigate their performance on predicting the transtibial prosthetic socket shape based on a 3D scan of the residual limb. While standardization is not required for all methods, it helps to ensure a fair comparison between them.

The first method uses PointNet$++$ layers \citep{qi_2017, qi_2017_2}, which are developed to learn directly from point clouds/meshes, managing challenges such as unstructured data. In this study, PointNet$++$ layers are implemented in a regression neural network predicting the transtibial socket shape. The second method, a Feedforward neural network, equips its architecture to learn from either 1) the standardized representation or 2) a dimensionality-reduced representation, made using principal component analysis (PCA) of the 3D meshes. The third method fits a random forest, an ensemble learning algorithm, on either 1) the standardized representation or 2) a PCA representation of the 3D meshes. 

\section{Methods}

\subsection{Data}
The dataset used in this study was collected from people with a transtibial amputation, each treated by a prosthetist working in the Dutch healthcare system \citep{prosthetic_dataset2024}. Each data sample consisted of a paired 3D scan of a patient's residual limb and the corresponding 3D model of the transtibial socket designed by the treating prosthetist. In addition to the 3D information, incomplete characteristics, such as socket type (PTB, TSB), patient's weight, and time since amputation of the patients, were also available. It is known from previous research that such characteristics influence the socket design \citep{radcliffe_1961, staats_1987, li_2019, dickinson_2021, dickinson2024}. However, given the incompleteness of these characteristics in the dataset, this research only used the 3D data to predict the transtibial prosthetic socket shape using multiple algorithms. 

\subsubsection{Pre-processing}
\begin{figure}[ht]
    \centering
    \includegraphics[width=0.8\linewidth]{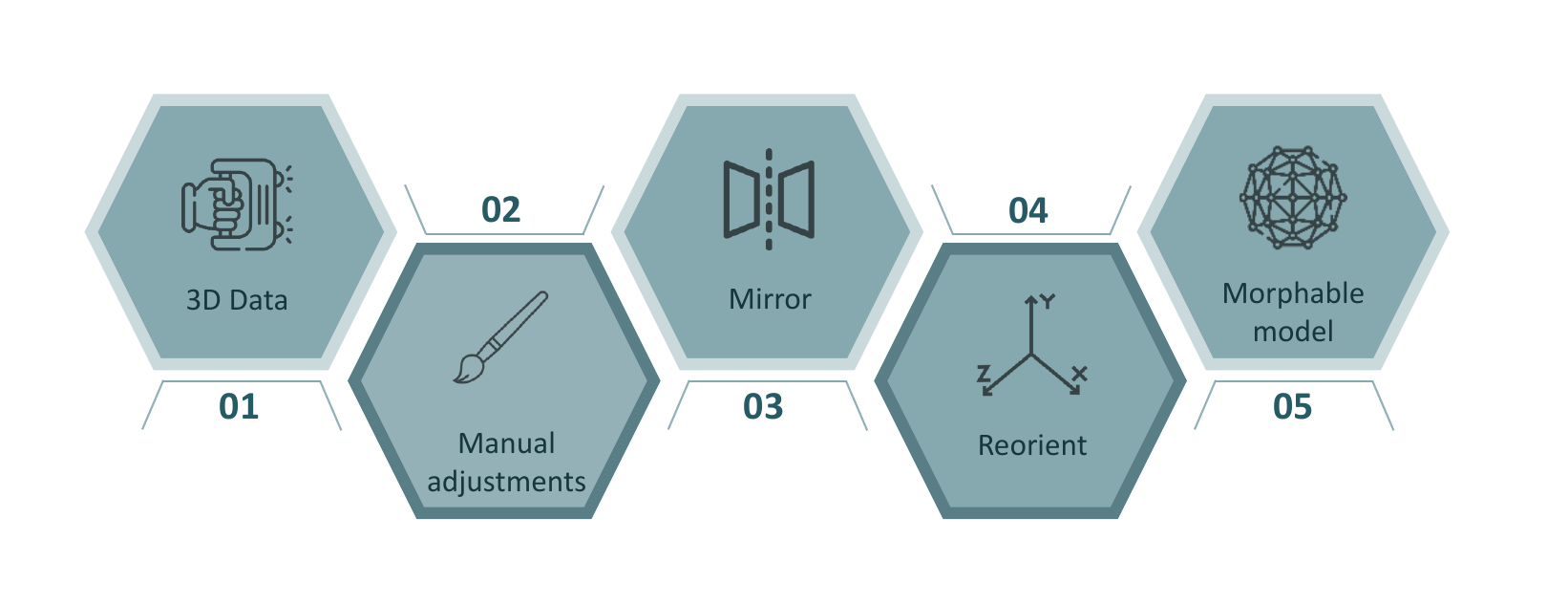}
    \caption{The five-step digital pre-processing workflow, consisting of initial data, manual adjustments, mirroring of right residual limbs, reorientation, and application of a morphable model.}
    \label{fig:pre_processing_workflow}
\end{figure}

To ensure that the 3D scans of the residual limb and corresponding 3D socket models can be used for training, multiple pre-processing steps were performed to prepare the data. This pre-processing workflow consists of five steps, as shown in Figure \ref{fig:pre_processing_workflow}. 

After loading the 3D data, artifacts were manually assessed on the 3D scans and models. Minor artifacts, such as small holes or uneven surfaces/spikes, inherent to these 3D scans, are corrected by filling or smoothing the scans in Meshmixer software version 3.5 (Autodesk Research, San Rafael, Calif). Scans incorrectly aligned or including larger holes located in regions of particular interest or containing the designated trimline area were excluded from the dataset. The dataset contains data from patients with left and right transtibial amputations. To ensure that the amputation side would not become a learnable feature, all right stumps were mirrored to represent them as left stumps. This side was chosen arbitrarily.

The data was reoriented to enforce that all scans and models were in the same orientation and space \citep{python}. This reorientation depends on two manually identified landmarks, the mid-patella and the end of the tibia. Since the 3D scans and corresponding 3D models are already matched, these landmarks are only identified on the 3D scans. Three technical physicians annotated these landmarks for all scans using 3DMedX\textsuperscript{\textregistered} software Version 1.2.35.0 (3D Lab Radboudumc, Nijmegen, The Netherlands). The inter-observer variability per landmark was measured to show the variability. Given that landmark determination on 3D scans is subject to inter-observer variability \citep{bermejo_2021}, variations of these annotated landmarks were used during development for algorithm robustness. Based on the variation in the annotated landmarks, 25 evenly distributed combinations of mid-patella and end of the tibia were generated for each patient using a Gaussian distribution. These orientations were used during network training, while the average of the landmarks from the three raters was used for testing.  Based on these two points, the required rotation in the x and y-plane could be calculated such that the clicked tibia point is on both the x and y-origin. The z-plane rotation is calculated using the mid-patella point and the weighted center of mass at this height. The models are translated such that the mid-patella is at the origin, and the tibia point is on both the x- and y-origin. An example can be found in \ref{appendix_preprocessing}, Figure \ref{fig:reorientation}.

As mentioned in the theoretical background, a morphable model, Meshmonk \citep{white_2019}, was applied to the scans and models (3DMedX\textsuperscript{\textregistered} software Version 1.2.36.2, 3D Lab Radboudumc, Nijmegen, The Netherlands) to ensure uniform model sampling. An average template was used for scan sampling, which was subsequently applied as a template for the socket models. This ensures that all 3D data consists of 3361 near-equally distributed vertices and 6672 faces.

\subsubsection{Principal component analysis}
PCA is applied to the data to investigate an alternative way of representing the scan stump pairs and remove noise present in the higher principal components. PCA was fitted on 1) the 3D scans, 2) the 3D socket models, and 3) the adaptations made by the prosthetist. Based on this reduced representation, the number of principal components explaining at least 95\% of variation was used as input and output representation. 

\subsection{Algorithm development}
\begin{figure}[ht]
    \centering
    \includegraphics[width=0.65\linewidth]{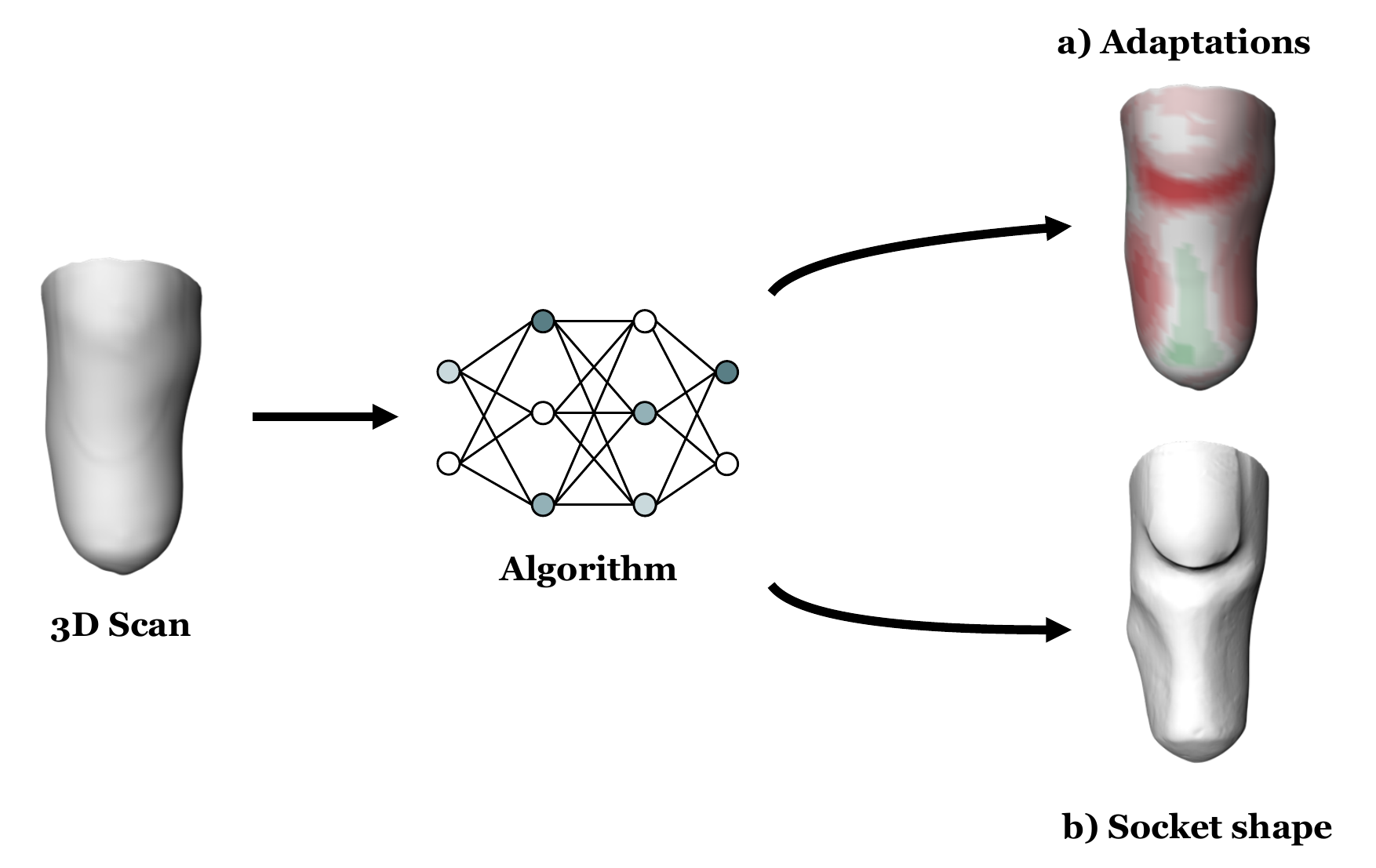}
    \caption{The algorithms use the 3D scan of the stump as input and predict either a) the adaptations, where red indicates reduction and green indicates enlargement of the areas, or b) the final socket shape}
    \label{fig:algorithm_output_options}
\end{figure}

Three different methods are tested and evaluated to investigate the capability of AI algorithms to predict the transtibial prosthetic socket shape based on a 3D scan of the residual limb. Two network implementations are evaluated for all methods, either to predict 1) the final socket shape or 2) the adaptations
performed by a prosthetist to predict the socket shape based on the 3D scan of the residual
limb (Figure \ref{fig:algorithm_output_options}). The adaptations are calculated by computing the difference between the original 3D scan and the final transtibial prosthetic socket shape. When these adaptations are applied to the original 3D scan,  the final transtibial prosthetic socket shape is constructed.

Consistently across methods, Pytorch was used for implementation, and the loss used was SmoothL1Loss \citep{Pytorch}. For the neural networks, a learning rate of $9.9 \times 10^{-4}$ was used in combination with an Adam Optimizer \citep{adam_optimizer}. Optimization of parameters and architectures was performed using Bayesian optimization from the Weights and Biases platform \citep{weights-biases-the-ai-developer-platform-2023}, minimizing the test loss using a set seed with 80\% of data for training and 20\% for testing. After optimization, 5-fold cross-validation with a set random seed was used to develop the final methods.

\subsubsection{Three-dimensional Neural Network}
\begin{figure}[H]

\sbox\twosubbox{%
  \resizebox{\dimexpr1\textwidth-1em}{!}{%
    \includegraphics[height=3cm]{example-image-a}%
    \includegraphics[height=3cm]{example-image-16x9}%
  }%
}
\setlength{\twosubht}{\ht\twosubbox}

\centering
\subcaptionbox{{Architectures for the neural networks with 3D Pointnet$++$ layers. Corresponding input and output sizes for the layers can be found in (b).  Regularization consists of batch normalization (momentum = 0.25) and dropout (rate = 0.25) \citep{qi_2017, yan_2019}}.}{%
  \includegraphics[height=\twosubht]{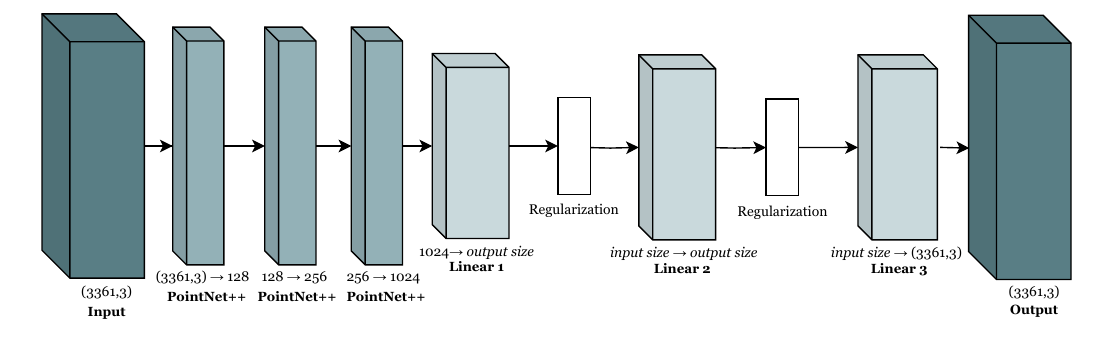}%
}\vspace{0.35cm}\quad
\hfill
\subcaptionbox{Input and output sizes for the layers in (a) given the predicted output.}{%

  \begin{adjustbox}{max width=0.5\textwidth}
        \begin{tabular}{|c|c|c|c|}
            \hline
            \textbf{Predicted Output} & \textbf{Layer} & \textbf{Input Size} & \textbf{Output Size} \\
            \hline
            \rowcolor[gray]{0.8} & Linear 1 & 1024 & 64 \\
             \rowcolor[gray]{0.8} & Linear 2 & 64 & 32 \\
             \rowcolor[gray]{0.8} \multirow{-3}{*}{Adaptations} & Linear 3 & 32 & (3361, 3)\\
            \hline
             & Linear 1 & 1024 & 2048 \\
              & Linear 2 & 2048 & 64 \\
             \multirow{-3}{*}{Socket Shape} & Linear 3 & 64 & (3361, 3)\\ 
             \hline
        \end{tabular}
        \end{adjustbox}
}
\caption{Architectures of 3D neural network}
\label{fig:architecture_PN}
\end{figure}

This method utilizes a neural network architecture containing PointNet$++$ layers by \citet{qi_2017}, to tackle the regression problem of predicting the transtibial prosthetic socket shape. As shown in Figure \ref{fig:algorithm_output_options}, the network is developed to predict either the socket shape or the required adaptations. Figure \ref{fig:architecture_PN} shows the architectures used to train these networks. The architectures contain three PointNet$++$ layers, followed by three linear layers. The  PointNet$++$ Pytorch implementation by \citet{yan_2019} is used. After epoch optimization, the socket shape algorithm is trained for 70 epochs, and the adaptations network is trained for 65 epochs.
\newpage
\subsubsection{Feed Forward Neural network}
\begin{figure}[ht]

\sbox\twosubbox{%
  \resizebox{\dimexpr0.9\textwidth-1em}{!}{%
    \includegraphics[height=3cm]{example-image-a}%
    \includegraphics[height=3cm]{example-image-16x9}%
  }%
}
\setlength{\twosubht}{\ht\twosubbox}

\centering
\subcaptionbox{{Architectures for the neural networks. Corresponding input and output sizes for the layers can be found in (b). Regularization 1  consists of batch normalization (momentum = 0.25), and regularization 2 consists of batch normalization (momentum = 0.25) and dropout (rate = 0.25). }}{%
  \includegraphics[height=\twosubht]{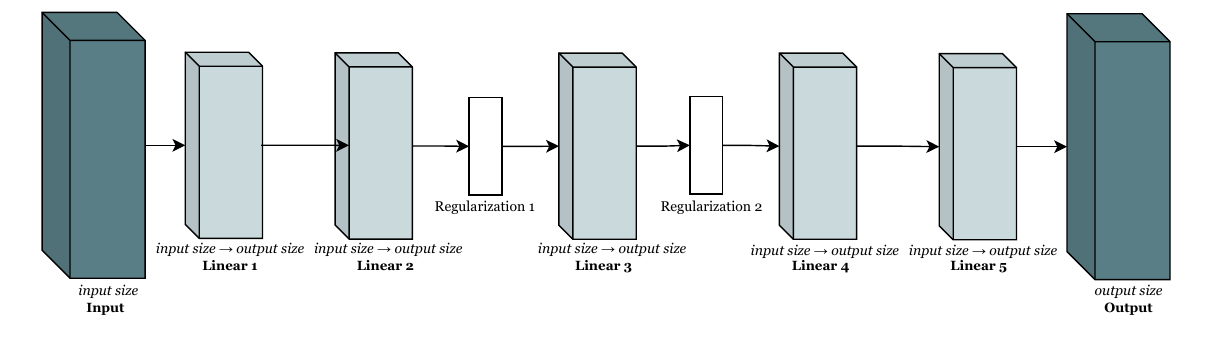}
}\vspace{0.35cm}\quad
\hfill
\subcaptionbox{Input and output sizes for the layers in (a) given the predicted output. For the output adaptations (reduced), only regularization 2 is used.}{%

  \begin{adjustbox}{max width=0.5\textwidth}
        \begin{tabular}{|c|c|c|c|c|}
            \hline
            \textbf{Predicted Output} & \textbf{Layer} & \textbf{Input Size} & \textbf{Output Size} \\
            \hline
            \rowcolor[gray]{0.8}  & Linear 1 & (3361,3) & 64 \\
            \rowcolor[gray]{0.8}  & Linear 2 & 64 & 128 \\
            \rowcolor[gray]{0.8}  & Linear 3 & 128 & 256\\ 
            \rowcolor[gray]{0.8}  & Linear 4 & 256 & 512\\ 
            \rowcolor[gray]{0.8} \multirow{-5}{*}{Adaptations} & Linear 5 & 512 & (3361,3) \\
            \hline
            \multirow{5}{*}{Adaptations (reduced)} & Linear 1 & 50 & 32 \\
             & Linear 2 & 32 & 32 \\
             & Linear 3 & - & - \\
             & Linear 4 & - & - \\
             & Linear 5 & 32 & 60 \\
             \hline
            \rowcolor[gray]{0.8}  & Linear 1 & (3361, 3) & 32 \\
            \rowcolor[gray]{0.8}  & Linear 2 & 32 & 512 \\
            \rowcolor[gray]{0.8}  & Linear 3 & 512 & 512\\ 
            \rowcolor[gray]{0.8}  & Linear 4 & 512 & 1048\\ 
            \rowcolor[gray]{0.8} \multirow{-5}{*}{Socket Shape} & Linear 5 & 1048 & (3361,3) \\
            \hline
            \multirow{5}{*}{Socket Shape (reduced)} & Linear 1 & 50 & 64 \\
             & Linear 2 & 64 & 128 \\
             & Linear 3 & 128 & 256\\
             & Linear 4 & 256 & 512 \\
             & Linear 5 & 512 & 70 \\
             \hline
        \end{tabular}
        \end{adjustbox}
}
\caption{Architectures of Feedforward neural network}
\label{fig:architecture_NN}
\end{figure}

The second method in this research uses a Feedforward neural network to predict either the socket shape or the required adaptations. Both outcomes are predicted using either 1) the vertices of the 3D scan and socket or vertex displacements of the adaptations or 2) the dimensionality-reduced version, using PCA, of these as input and output. Figure \ref{fig:architecture_NN} shows the used architectures for each output. After epoch optimization, the adaptations network is trained for 50 epochs, the reduced adaptations network is trained for 75 epochs, the socket shape network is trained for 175 epochs, and the reduced socket shape network is trained for 150 epochs.
\subsubsection{Random forest}
The third method fits a random forest regressor to predict the transtibial prosthetic socket shape or the necessary adaptations. Similar to the previous method, this method also investigates the use of a dimensionality-reduced version (PCA) of the data on the algorithms' performance. Table \ref{table:fitting_parameters_RF} shows the final parameters that fit the random forest.
\begin{table}[H]
\begin{adjustbox}{max width=1\textwidth}
\begin{tabular}{|c|c|c|c|c|c|c|c|}
    \hline
    \textbf{Predicted} & \textbf{Number} & \textbf{Maximum} & \textbf{Minimum} & \textbf{Minimum} & \textbf{Maximum} \\
    \textbf{output} & \textbf{of estimators} & \textbf{depth} & \textbf{samples split} & \textbf{samples leaf} & \textbf{features} \\
    \hline
    \rowcolor[gray]{0.8} Adaptations & 1154 & 162 & 6 & 3 & $sqrt$  \\
    \hline
    Adaptations (reduced) & 1297 & 179 & 5 & 1 & $sqrt$  \\
    \hline
    \rowcolor[gray]{0.8} Socket shape & 920 & 181 & 5 & 1 & $log_2$  \\
    \hline
    Socket shape (reduced) & 320 & 158 & 6 & 5 & $sqrt$  \\
    \hline 
\end{tabular}
\end{adjustbox}
\caption{Parameters used for fitting of random forest given the predicted output given the predicted output}
\label{table:fitting_parameters_RF}
\end{table}

\section{Experimental Setup}
We used multiple metrics to compare our methods. The first metric is the Euclidean distance that captures the three-dimensional difference between the predicted and true vertices of the mesh. This metric compares the displacement between the location of the predicted and target vertices. This metric is used to give an overall indication of the fit of the network and was used during training. However, a displacement between the locations of the vertices does not always indicate the same displacement between the predicted shapes formed by the vertices. For a more precise representation, the closest point from the predicted mesh to the target mesh, the surface-to-surface distance, was calculated for the test set. This metric gives a better representation of the differences between the shapes of the transtibial sockets. Lastly, distance maps between the predictions and the golden standard were utilized to analyze the errors' location.

\section{Results}
\subsection{Data}
Of the 202 available scan pairs, 84 were excluded due to either 1) incorrect alignment of the 3D scan of the stump compared to the 3D model of the socket or 2) artifacts on the scans, such as holes or trim blocks, used to decide where to cut the prosthesis. Resulting in a total of 118 scan pairs used for algorithm development.  

\begin{figure}[ht]
    \centering
    \includegraphics[width=0.7\linewidth, height = 5cm]{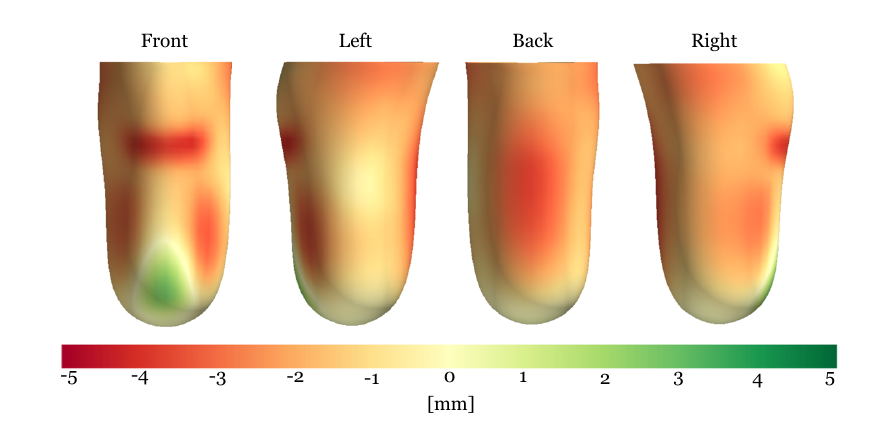}
    \caption[Average adjustments]{Average adjustments from stump to socket in millimeters. Red indicates reduction, and green indicates enlargement.}
    \label{fig:average_adjustments}
\end{figure}

The data is from 74 males, 27 females, and 18 unreported genders. Of these patients, 56 had a left amputation and 62 a right amputation. The remaining demography of the used data can be found in Table \ref{tab:demography_data}. To get an overall idea of the dataset and a simplified view of the adaptations required to create a fitting socket, the average adaptations to create a socket from a scan are shown in Figure \ref{fig:average_adjustments}.

\begin{table} [ht]
    \centering
    \begin{tabular}{|c|c|c|}
    \hline
    Characteristic & Average (std) & Unknown values \\
    \hline
    \rowcolor[gray]{0.8} Age [years] & 64 (17) & 19 \\
    Body length [cm] & 174.9 (9.6) & 51 \\
    \rowcolor[gray]{0.8} Body weight [kg] & 79.9 (15)& 50 \\
    Time since amputation [years] &  5.5 (2.7) & 62 \\
    \rowcolor[gray]{0.8} Stump Circumference [cm] & 34.9 (3.9) & 0 \\ 
     Stump length [cm] & 20.9 (2.8) & 0 \\
    \hline
    \end{tabular}
    \caption{Data demography, showing the average, standard deviation and number of unknown values for age, body length and weight, time since amputation, stump length and circumference.}
    \label{tab:demography_data}
\end{table}

\subsection{Landmarking}
As mentioned in the data pre-processing, the reorientation was performed based on two landmarks: the mid-patella and the tibia-end. The statistics of the average error and inter-observer variability between the three annotators are shown in Table \ref{tab:landmark_error_statistics}. 

\begin{table}[H]
\centering
\begin{adjustbox}{max width=\textwidth}
\begin{tabular}{c|c|c|c|c|}
 \cline{2-5}
\centering

 & \multicolumn{4}{|c|}{\textbf{Inter-oberserver variability [mm]}} \\
 \cline{2-5}
 & \multicolumn{2}{|c|}{Mid-patella} & \multicolumn{2}{|c|}{Tibia-end} \\
 \cline{2-5}
 & Median& (Q1, Q3) & Median& (Q1,Q3) \\
 \hline
 \multicolumn{1}{|c|}{ \cellcolor[gray]{0.8} Average}  &  \cellcolor[gray]{0.8}7.48& \cellcolor[gray]{0.8} (5.43, 11.56)& \cellcolor[gray]{0.8}10.34& \cellcolor[gray]{0.8}(7.32, 14.95) \\
 \hline
 \multicolumn{1}{|c|}{Observer 1 vs 2} &  8.13 & (5.05, 11.42) & 9.81 & (6.23, 16.49)\\
 \hline
 \multicolumn{1}{|c|}{\cellcolor[gray]{0.8}Observer 2 vs 3} & \cellcolor[gray]{0.8}6.53 & \cellcolor[gray]{0.8} (3.56, 11.35) & \cellcolor[gray]{0.8}9.29 & \cellcolor[gray]{0.8}(5.72, 14.07)\\
  \hline
 \multicolumn{1}{|c|}{Observer 1 vs 3} &  7.28 & (4.44, 10.89) & 10.43& (6.42, 16.62)\\
\hline
\end{tabular}
\end{adjustbox}
\caption{Average errors and inter-observer variability of landmark identification in millimeters.}
\label{tab:landmark_error_statistics}
\end{table}

\subsection{Algorithms}
Table \ref{tab:comparison_methods_average} shows the test set's corresponding surface-to-surface distances. For all methods, training to learn the necessary adaptations yields better results than training to predict the final transtibial prosthetic socket shape. The random forest that predicts the non-reduced adaptations performs best with a mean Euclidean distance of 1.86 millimeters, a median surface-to-surface distance of 1.24 millimeters, a first quartile of 1.03 millimeters, and a third quartile of 1.54 millimeters over the test set. 

\begin{table}[H]
\centering
\begin{adjustbox}{width=0.8\textwidth, height = 2.5cm}
\begin{tabular}{|c|c|c|c|}
\hline
&  & \multicolumn{2}{|c|}{\textbf{Surface-to-surface [mm]}}  \\
\cline{3-4}
\multirow{-2}{*}{\textbf{Predicted Output}} & \multirow{-2}{*}{\textbf{ Method}} & Median & (Q1, Q3) \\ 
\hline
\rowcolor[gray]{0.8} & 3D Neural Network & 1.31 & (1.16, 1.62) \\
 \rowcolor[gray]{0.8} & Neural Network & 1.27 & (1.13, 1.57) \\
 \rowcolor[gray]{0.8} \multirow{-3}{*}{Adaptations}& Random Forest & \textbf{1.24} & \textbf{(1.03, 1.54)} \\
 \hline
& Neural Network & 1.27 & (1.10, 1.52)  \\
 \multirow{-2}{*}{Adaptations (reduced)} & Random Forest & 1.29 & (1.14, 1.62) \\
\hline
\rowcolor[gray]{0.8} & 3D Neural Network & 6.10 & (5.02, 8.73) \\
\rowcolor[gray]{0.8} & Neural Network & 4.90 & (3.64, 6.59) \\
\rowcolor[gray]{0.8} \multirow{-3}{*}{Socket shape} & Random Forest & 3.84 & (3.12, 4.84) \\
\hline 
\multirow{2}{*}{Socket shape (reduced)} & Neural Network & 5.00 & (3.96, 6.57) \\
& Random Forest & 5.84 &  (4.81, 7.96) \\ 
\hline
\end{tabular}
\end{adjustbox}
\caption{The test surface-to-surface distance averaged over the five folds}
\label{tab:comparison_methods_average}
\end{table}

\begin{figure}[H]
    \centering
    \includegraphics[width=0.8\linewidth, height = 5.5cm]{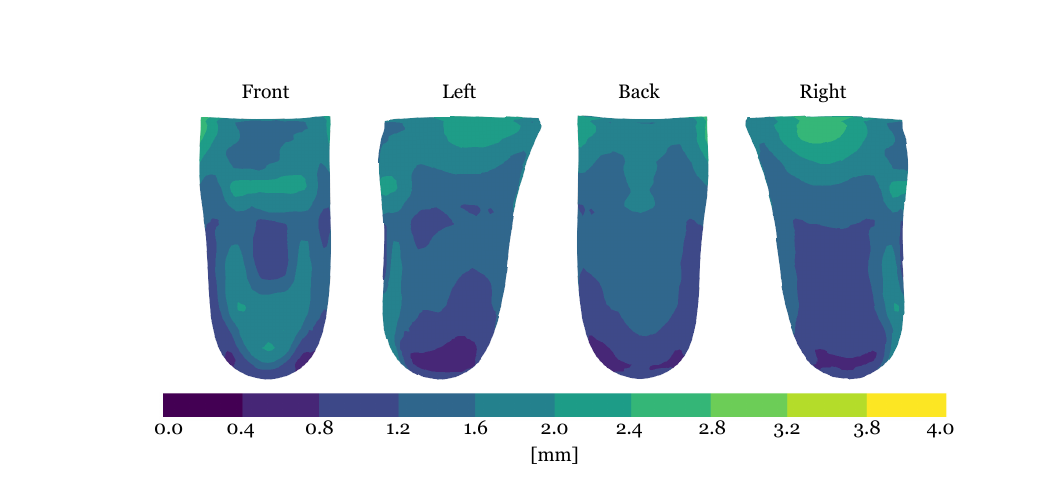}
    \caption{The absolute location of the surface-to-surface error for the random forest predicting the adaptations in millimeters}
    \label{fig:sts_distancemap}
\end{figure}

\begin{figure}[H]
    \centering
    \begin{subfigure}{0.49\linewidth}
        \centering
        \includegraphics[width=\linewidth]{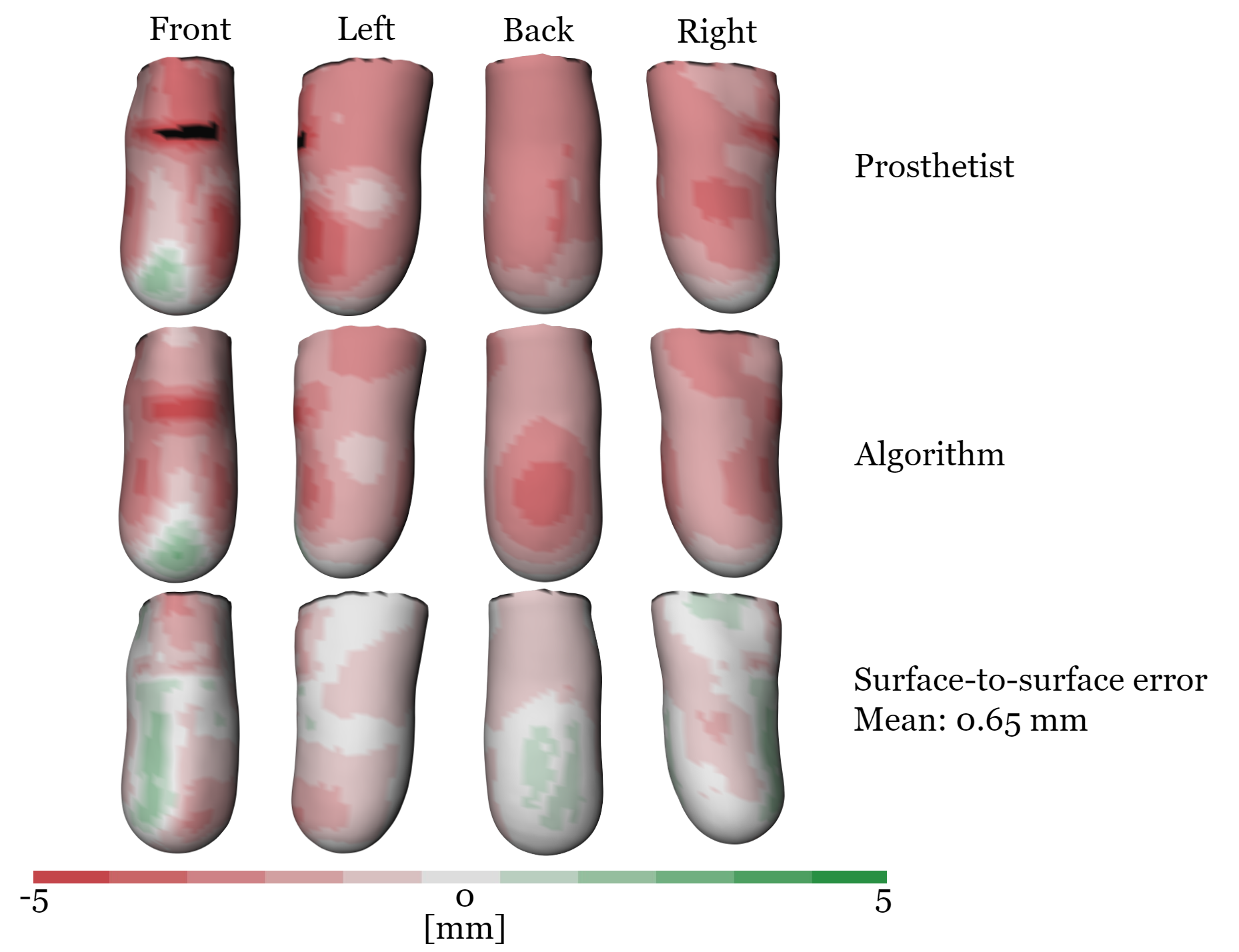}
        \caption{Best test case}
        \label{fig:best_distancemap}
    \end{subfigure}
    \hfill
    \begin{subfigure}{0.49\linewidth}
        \centering
        \includegraphics[width=\linewidth]{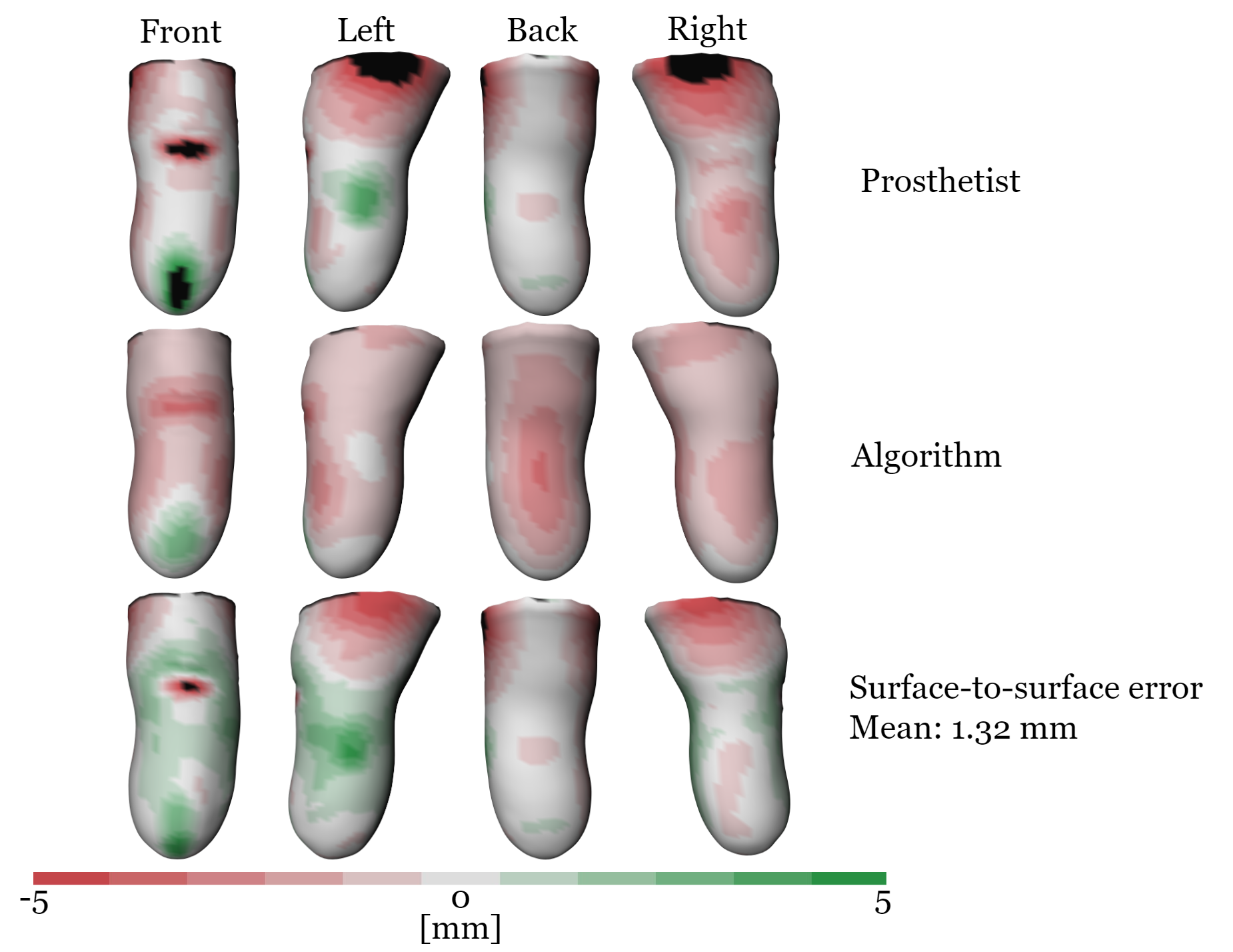}
        \caption{Median test case}
        \label{fig:median_distancemap}
    \end{subfigure}
    \hfill
    \begin{subfigure}{0.49\linewidth}
        \centering
        \includegraphics[width=\linewidth]{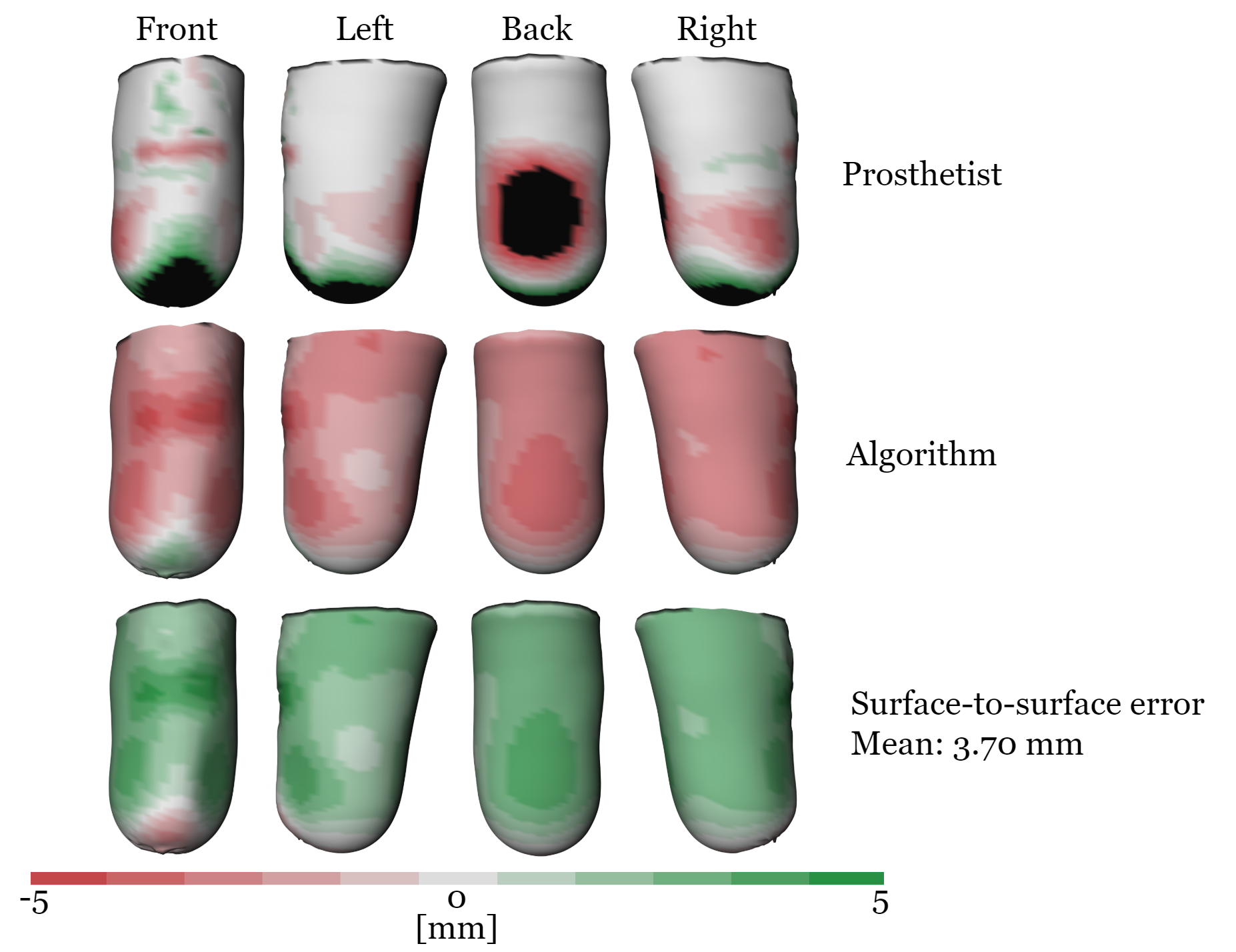}
        \caption{Worst test case}
        \label{fig:worst_distancemap}
    \end{subfigure}
    \caption{Distance maps of the surface-to-surface error for best, median, and worst test case. Shown for the adaptations made by the prosthetist, the algorithm, and the difference between the algorithm and the prosthetist. For visibility, black indicates an adaptation or difference that is out of the scale range.}
    \label{fig:distance_maps}
\end{figure}

Figure \ref{fig:sts_distancemap} shows the location of the absolute surface-to-surface error for this method. Figure \ref{fig:distance_maps} shows the distance maps for the networks' best, median, and worst predicted sockets. It is seen that the algorithm adapts its predictions based on the input; more intense reduction is given to bigger stumps with a bigger circumference (Figure \ref{fig:worst_distancemap}). In comparison, longer stumps show more evenly distributed adaptations (Figure \ref{fig:median_distancemap}). Looking at the adaptations performed by the prosthetists, it can be seen that the average adaptations (Figure \ref{fig:average_adjustments}) shown in this paper do not capture specific transtibial prosthetic socket shape cases. Looking at the specific cases shown in the results (Figure \ref{fig:distance_maps}), it can be seen that the adaptations made by a prosthetist vary from local regions with or without whole volume changes.    

\newpage
\section{Discussion}
This study investigated multiple methods to standardize the creation of the transtibial prosthetic socket shape. The results demonstrated that, when predicting the transtibial prosthetic socket shape from a scan, estimating the adaptations yields better performance than directly predicting the socket shape, across all tested methods. 

We hypothesize that the amount of data available in this study limits the performance of these networks. An increase in available data may help reduce the performance gap between predicting the adaptations and predicting the final socket shape. The results also show that the network using PointNet$++$ layers does not yield better results when compared to the other less computationally expensive methods \citep{qi_2017}. Looking at the use of principal components as an abstraction for the adaptations, it can be seen that similar results were reached for the tested methods compared to directly using the 3D data. However, given that not all principal components were used (minimum 95\% explained variance) for the reduced representation of the data, a reconstruction error occurs when the original data is reproduced using these principal components. Given that this reconstruction error is not present when using the 3D data directly, it is promising that similar results occur. This could indicate that methods removing 'noise' can still capture the necessary information. 

Comparing our results to previous literature, this study showed that PCA can not only describe the stump and socket shape \citep{dickinson_2021} but can also be used to predict a transtibial prosthetic socket shape when used in an algorithm. These findings are consistent with the research by \citet{cutti_2024}, which used PCA to predict the socket shape of one prosthetist. However, we showed that such an approach can be extended to a larger dataset, including multiple prosthetists and different types of sockets. Giving the performance of a mean Euclidean distance of 2.51 millimeters, found by \citet{van_der_stelt_2024}, our algorithm's performance of a mean Euclidean distance of 1.86 millimeters outperforms theirs. Given that the sockets created by the algorithm from \citet{van_der_stelt_2024} fitted satisfactorily on eight out of ten patients, we expect to reach similar or better results given our improvements on the test set. In the current literature, there is a gap in knowledge about the inter- and intra-observer variability in the rectifications made to create a socket shape. \citet{dickinson_2022} did quantify the differences between capturing the residual limb shape using plaster casts and using different 3D scanners, but this does not quantify the differences between prosthetists when creating socket shapes. Therefore, our results cannot be directly generalized to acceptable deviation in socket shape and quality.  

This research has multiple limitations. First, the landmarks were not annotated by the prosthetist but were placed retrospectively on the 3D scan, which resulted in interobserver variability. Given that these landmarks are used for the reorientation of the scan pairs, variation in these annotations results in variations in the orientation. To counter this, this research used multiple orientations of scans and corresponding socket models, based on the variation in annotations, were used during training. Second,  for a fair comparison of the investigated algorithms, the approaches were standardized. For a real-world application of one of these algorithms, it is advised to perform data- and algorithm-specific optimization to get the best possible performance. Last, it was known that the used dataset is heterogeneous since it is, for example, based on multiple prosthetists, different kinds of sockets (PTB, TSB), and different amounts of time that passed since amputation. Currently, such characteristics are not incorporated in the algorithms' predictions, but do influence the socket design \citep{radcliffe_1961, staats_1987, li_2019, dickinson_2021, dickinson2024}.

Given the aforementioned results and limitations of this research, multiple points for future work can be addressed. First, the variations in landmarks annotations on the 3D scans \citep{bermejo_2021} could be addressed by letting the prosthetist place landmarks on the stump before scanning. We expect this to improve accuracy, given that the ability to feel the residual limb gives more insights into anatomical landmarks. Second, the heterogeneous dataset leaves room for improvement by incorporating characteristics, such as socket type, to predict the transtibial prosthetic socket shape. Future research will focus on expanding the dataset to include additional variables, such as socket type, weight of the patient, and time since amputation. These complementary features can be integrated with the 3D data to train more advanced multimodal algorithms. Last, given that the surface-to-surface errors are higher in patient-specific areas, it would be interesting to research custom loss functions that emphasize attention to these areas. 
    
\section{Conclusion}
This study compared the use of multiple algorithms and data representations to predict the interior wall of the prosthetic transtibial socket shape. Predicting the necessary adaptations yields better results than predicting the final socket shape. Furthermore, it was shown that using PCA to predict the socket shape did not substantially improve the results. The random forest predicting the necessary adaptations performed best, yielding a median surface-to-surface distance of 1.24 millimeters, a first quartile of 1.03 millimeters, and a third quartile of 1.54 millimeters. These results are promising for the development of algorithms aimed at standardizing transtibial prosthetic socket design in 
LMICs.

\newpage 
\section*{Aknowledgements}
\noindent We would like to thank OIM Orthopedie (Assen, the Netherlands) for providing the data and StitPro and ZonMw for co-financing the AI-development. Funding: This work was supported by StitPro; ZonMw program: Goed Gebruik Hulpmiddelen Thuis, project number:10310012110003.

\bibliographystyle{elsarticle-num-names} 
\bibliography{bibliography}

\appendix
\section{Pre-processing} \label{appendix_preprocessing}
\begin{figure}[H]
    \centering
    \includegraphics[width=0.75\linewidth]{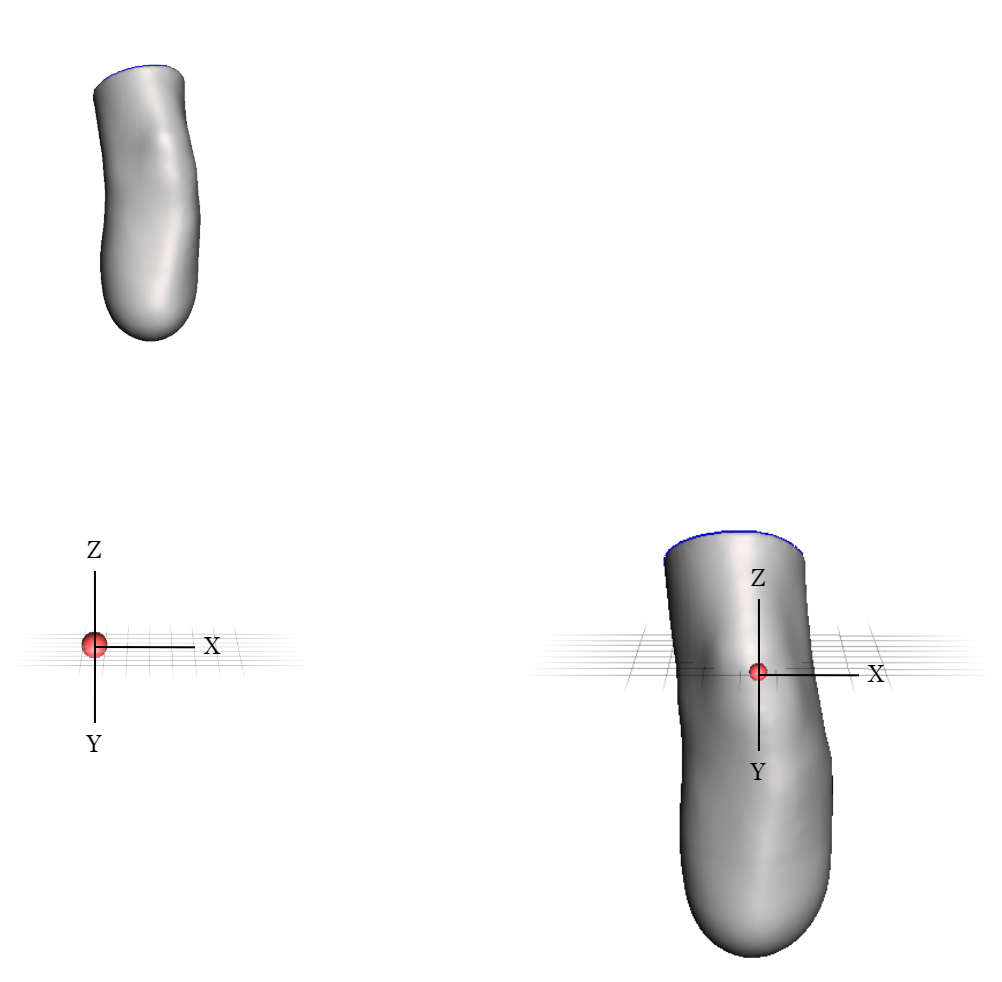}
    \caption{An example of a 3D scan before reorientation (left) and after reorientation (right). During reorientation the scan is translated and rotated such that the mid-patella is at the origin (0,0,0) and the tibia point is on both the x and y-origin. 
    }
    \label{fig:reorientation}
\end{figure}

\end{document}